\documentclass{article}

\usepackage{microtype}
\usepackage[table,dvipsnames]{xcolor}
\usepackage{colortbl}
\usepackage{graphicx}
\usepackage{subcaption}
\usepackage{booktabs} % for professional tables
\usepackage{amsmath,amssymb} % define this before the line numbering.
\usepackage{color}

\usepackage{booktabs}
\usepackage{multirow}

\usepackage{hyperref}

\usepackage[accepted]{icml2025}

\usepackage{amsmath}
\usepackage{amssymb}
\usepackage{mathtools}
\usepackage{amsthm}

\usepackage{pifont}% http://ctan.org/pkg/pifont
\usepackage{enumitem}
\let\oldding\ding% Store old \ding in \oldding
\renewcommand{\ding}[2][1]{\scalebox{#1}{\oldding{#2}}}

\usepackage[capitalize,noabbrev]{cleveref}

\theoremstyle{plain}

\theoremstyle{definition}

\theoremstyle{remark}

\usepackage[textsize=tiny]{todonotes}

\newcommand{\dlfr}[0]{\texttt{DLFR-VAE}}
\newcommand{\syz}[1]{\textcolor{blue}{SYZ:}}

\icmltitlerunning{\dlfr: Dynamic Latent Frame Rate VAE for Video Generation}

\begin{document}

\twocolumn[
\icmltitle{\dlfr: Dynamic Latent Frame Rate VAE for Video Generation}



\icmlsetsymbol{equal}{*}

\begin{icmlauthorlist}
\icmlauthor{Zhihang Yuan}{equal,yyy,comp}
\icmlauthor{Siyuan Wang}{equal,yyy}
\icmlauthor{Rui Xie}{yyy,comp}
\icmlauthor{Hanling Zhang}{yyy,comp}
\icmlauthor{Tongcheng Fang}{yyy,comp}
\icmlauthor{Yuzhang Shang}{iit}
\icmlauthor{Shengen Yan}{comp}
\icmlauthor{Guohao Dai}{shjt,comp}
\icmlauthor{Yu Wang}{yyy}
\end{icmlauthorlist}

\icmlaffiliation{yyy}{Tsinghua University}
\icmlaffiliation{comp}{Infinigence AI}
\icmlaffiliation{iit}{Illinois Tech}
\icmlaffiliation{shjt}{Shanghai Jiao Tong University}

\icmlcorrespondingauthor{Yu Wang}{yu-wang@tsinghua.edu.cn}
\icmlcorrespondingauthor{Yuzhang Shang}{yshang4@hawk.iit.edu}

\icmlkeywords{Machine Learning, ICML}

\vskip 0.3in
]



% \icmlEqualContribution

\renewcommand{\thefootnote}{\fnsymbol{footnote}}
\footnotetext[1]{Equal contribution. $^1$Tsinghua University $^2$Infinigence AI
 $^3$Illinois Tech $^4$Shanghai Jiao Tong University.}
\renewcommand{\thefootnote}{\arabic{footnote}}

\begin{abstract}
In this paper, we propose the Dynamic Latent Frame Rate VAE (\textbf{\dlfr}), a training-free paradigm that can make use of adaptive temporal compression in latent space.
While existing video generative models apply fixed compression rates via pretrained VAE, we observe that real-world video content exhibits substantial temporal non-uniformity, with high-motion segments containing more information than static scenes. 
Based on this insight, \dlfr~dynamically adjusts the latent frame rate according to the content complexity.
Specifically, \dlfr~comprises two core innovations: \raisebox{-1.1pt}{\ding[1.1]{182\relax}} a Dynamic Latent Frame Rate Scheduler that partitions videos into temporal chunks and adaptively determines optimal frame rates based on information-theoretic content complexity, and
\raisebox{-1.1pt}{\ding[1.1]{183\relax}} a training-free adaptation mechanism that transform pretrained VAE architectures to dynamic VAE that can process features with variable frame rates.
Our simple but effective \dlfr~can function as a plug-and-play module, seamlessly integrating with existing video generation models and accelerating the video generation process. 
% Project link: \href{https://github.com/thu-nics/DLFR-VAE}{github.com/thu-nics/DLFR-VAE}.
% Experiments on UCF-101 and Kinetics-400 demonstrate that DLFR-VAE maintains high reconstruction fidelity, while reducing latent space elements by xxx\%, resulting in zz× faster inference and yy× lower memory usage in downstream generative models.
\end{abstract}

\section{Introduction}

Video is a fundamental medium for capturing real-world dynamics, making the generation of diverse video content a crucial capability for AI systems~\cite{videoworldsimulators2024,agarwal2025cosmos}. 
Recent advances in diffusion models~\cite{ho2020denoising,rombach2022high,esser2024scaling,jin2024pyramidal} and autoregressive models~\cite{kuaishou2024,fan2024fluid} have led to notable breakthroughs in producing high-fidelity and long-duration videos.

At the heart of these video generation frameworks lies the Variational Autoencoder (VAE) \cite{kingma2013auto,rombach2022high,xing2024large}, which jointly reduces spatial and temporal dimensions to create compact latent representations. 
This latent space not only provides a more structured manifold for downstream generative tasks, but also substantially lowers computational and memory requirements compared to operating in the original video domain~\cite{agarwal2025cosmos,videoworldsimulators2024}. Based on the latent space, modern architectures, such as Diffusion Transformers~\cite{peebles2023scalable} or Auto-Regressive Transformers~\cite{esser2021taming}, can then effectively learn the distribution of these latent representations, with the VAE decoder ultimately reconstructing them back into complete videos \cite{jin2024pyramidal,kuaishou2024,kong2024hunyuanvideo}.

The computational complexity of video generation is largely determined by the size of the latent representation (\textit{i.e.}, the number of latent tokens)~\cite{kong2024hunyuanvideo,zheng2024open,kondratyuk2023videopoet}. 
% . For a typical eight-second video processed at 24 frames per second with an 8$\times$ spatial compression ratio, the model must generate 10$\times$24$\times$512 = 122,880 tokens in the latent space
This substantial token count imposes significant computational overhead, primarily due to the quadratic complexity of attention in diffusion and autoregressive Transformers~\cite{vaswani2017attention,kondratyuk2023videopoet,peebles2023scalable}. Therefore, addressing this bottleneck is pivotal for extending video generation to longer durations and higher resolutions. 

\begin{figure}[!t]
    \centering
    \includegraphics[width=0.99\linewidth]{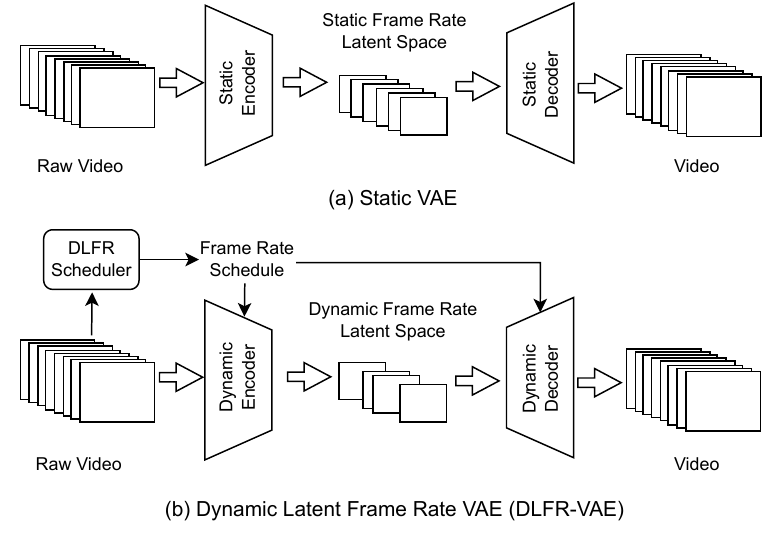}    
    \vspace{-0.2in}
    \caption{\textbf{\dlfr}: A training-free approach that accelerates video generation through content-adaptive spatial-temporal compression. This module can seamlessly integrate with existing pretrained video generative models.}
    % (both diffusion- and autoregressive-based) while preserving their generative ability.}
    \vspace{-0.3in}
    \label{fig:intro}
\end{figure}

In order to achieve more efficient video generation in a training-free manner, we explore the possibility of adjusting latent frame rates based on information-theoretic content density.
Prior studies have demonstrated that video data exhibits significant temporal non-uniformity~\cite{feichtenhofer2019slowfast,yang2020temporal,li2022nuta}. For example, high-motion segments (e.g., soccer shots) have more content complexity than static scenes (e.g., stationary players). 
% Overcompressing sparse-content segments causes information loss, whereas undercompressing dense-content segments causes redundancy.
Our empirical observations reveal that in standard fixed-rate latent spaces, roughly 35\% of latent units carry minimal information, indicating considerable computational waste. 
These findings support our key motivation: sparse-content segments (e.g., slow-motion sequences) can be efficiently represented with fewer latent elements, whereas dense-content segments require a higher token density to preserve temporal complexity~\cite{yu2024efficient,xiang2020zooming,lin2023spvos,ravanbakhsh2024deep}.

Based on this motivation, we propose \textbf{Dynamic Latent Frame Rate VAE, \dlfr}, a novel framework that enables dynamic frame rate in the latent space without additional training. 
We proposes two key technical innovations for \dlfr: \raisebox{-1.1pt}{\ding[1.1]{182\relax}} a Dynamic Latent Frame Rate Scheduler that partitions videos into temporal chunks and adaptively determines optimal frame rates based on content complexity, and \raisebox{-1.1pt}{\ding[1.1]{183\relax}} a training-free adaptation mechanism that enables pretrained VAE architectures to process features with variable frame rates. 
This dynamic frame rate strategy can significantly reduce both computational overhead and memory requirements for video generative models (See Fig.\ref{fig:intro}).

The first component \raisebox{-1.1pt}{\ding[1.1]{182\relax}} centers on our efficient content complexity metric based on inter-frame differences, which enables real-time frame rate adaptation during inference with minimal computational overhead.
We can then schedule each video chunk's adaptive frame rate based on this metric. 
Building on this scheduler, we develop \raisebox{-1.1pt}{\ding[1.1]{183\relax}} a training-free mechanism to augment pretrained VAE models with dynamic frame rate capabilities. Leveraging the robust compression and reconstruction capabilities of large-scale pretrained video VAEs~\cite{wu2024improved,xing2024large,chen2024deep}, we strategically inject dynamic sampling modules into the network architecture. 
In the encoder, we incorporate dynamic downsampling operators that modulates frame rates, producing chunk features with varying temporal resolutions. 
% These variable-rate features are processed through the remaining encoder layers to produce the dynamic latent representation. 
The decoder mirrors this design with dynamic upsampling operators that restores features to a consistent frame rate.
% based on encoding-time schedules. 
This architectural approach preserves pretrained weights while enabling dynamic frame rate control, making \dlfr~a plug-and-play module for downstream video generative models.
% whether under diffusion- or autoregressive-based~\cite{zheng2024open,kondratyuk2023videopoet} frameworks.

Experimental results show that DLFR maintains a reconstruction quality (SSIM change $<$ 0.03) while reducing the average number of latent space elements by 50\%. We integrated DLFR into an existing diffusion-based video generation model without any fine-tuning, and the model was able to generate acceptable videos with a significant speedup. The latency of a diffusion step is reduced by 2 to 6 times, and the end to end speedup is 2x to generate a video.

\section{Related Work}
\subsection{Autoencoders for Visual Generation Models} 
Visual generation in high-resolution pixel space imposes prohibitive computational costs for diffusion and autoregressive models. To address this, \citet{rombach2022high} introduced latent diffusion models operating in compressed spaces via pretrained autoencoders~\cite{kingma2013auto}. The standard design employs an $8\times$ spatial compression ratio with four latent channels \citep{peebles2023scalable,li2024autoregressive,tian2024visual}. Recent works have focused on improving reconstruction quality through increased latent channels \citep{esser2024scaling} or enhanced decoders with task-specific priors \citep{zhu2023designing}.

In contrast, our work targets a different but equally important goal: dynamically increasing the spatial compression ratio of autoencoders while still maintaining acceptable reconstruction quality. \citet{chen2024deep} also address high compression ratios to enable efficient high-resolution diffusion models; however, their approach trains a specialized autoencoder. Our method, by comparison, is training-free, allowing us to obtain a more compressed latent space without retraining the original autoencoder. To our knowledge, this is the first study to explore higher compression ratios in this training-free manner. 

\begin{figure*}[!th]
    \centering
    \begin{minipage}{\textwidth}
    \centering
    \begin{subfigure}{0.57\textwidth}
    \includegraphics[width=\linewidth]{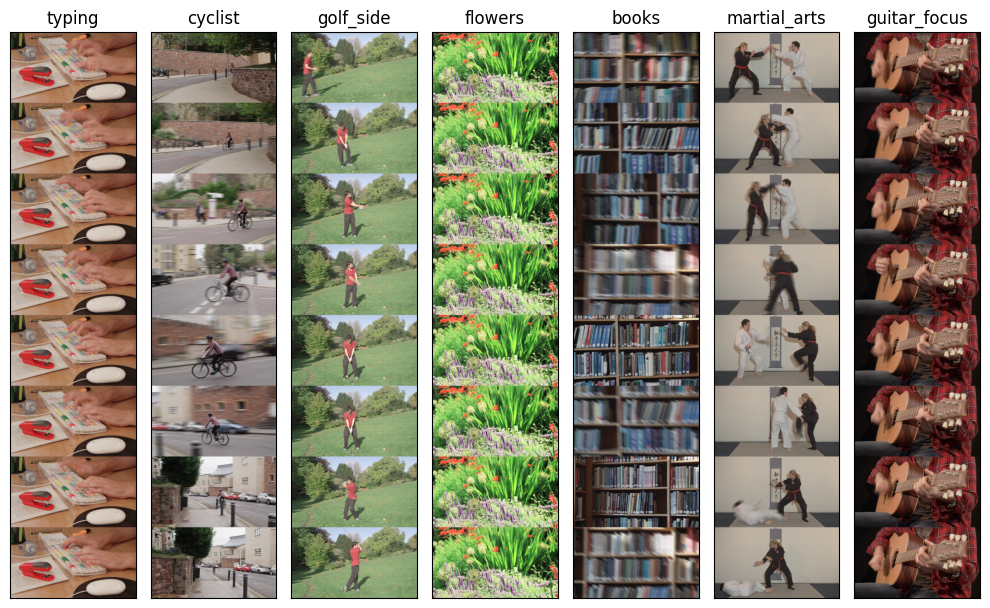}
    \captionsetup{font=small}
    \caption{Example video segments in BVI-HFR~\cite{mackin2018study}. Each one is 10 seconds at 1080p/60Hz. The ``cyclist'' and ``books'' demonstrates rapid camera motion, while ``flowers'' and ``typing'' contain slower, more static content.}
    \label{fig:video_examples}
    \end{subfigure}
    \hspace{0.05in}
    \begin{subfigure}{0.41\textwidth}
    \centering
    \includegraphics[width=0.92\linewidth]{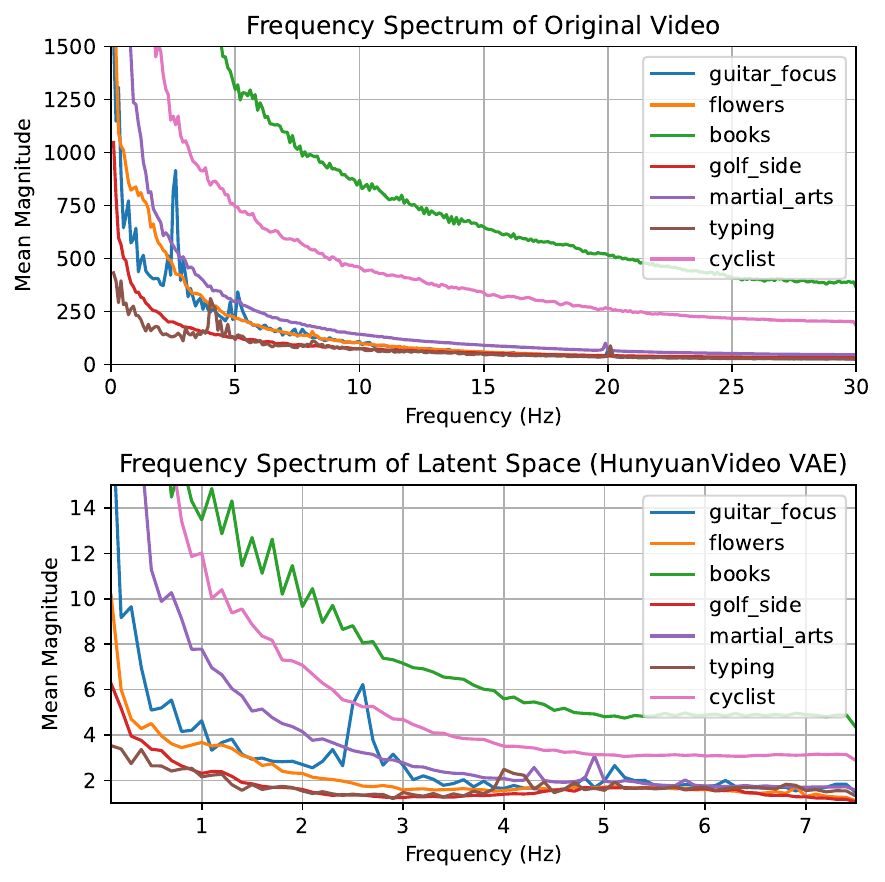}
    \captionsetup{font=small}
    \vspace{-0.1in}
    \caption{Temporal frequency analysis comparing original video signals (top) and their latent space representations from HunyuanVideo VAE (bottom) \cite{kong2024hunyuanvideo}.}
    \label{fig:analysis_on_video_seg}
    \end{subfigure}
    \end{minipage}
    \vspace{-0.13in}
    \caption{\textbf{Analysis of temporal frequency characteristics in both pixel and latent spaces.} Key observations: (1) Fast-motion segments exhibit higher temporal frequency content in both domains, while static scenes show concentrated low frequency. (2) The latent space preserves the relative frequency patterns of the original signals, enabling content-adaptive frame rate compression in the latent domain.}
    \vspace{-0.2in}
\label{fig:motivation}
\end{figure*}

\subsection{Training-free Acceleration for Generative Models}
The computational intensity of generative models has spurred various acceleration strategies, with training-free methods gaining prominence due to the high training costs of modern architectures~\cite{ma2024efficient}. Key approaches include: (1) Reducing inference sampling steps through training-free few-step samplers~\cite{song2020denoising,lu2022dpm,lu2022dpmp,zhang2022fast};
(2) Model compression via sparsity~\cite{ma2024deepcache,yuan2024ditfastattn} or quantization~\cite{shang2023post,li2023q,wang2024quest,wu2024ptq4dit,li2024svdqunat}.

While these methods focus on optimizing diffusion or autoregressive backbones, they leave the autoencoder unchanged. Our approach introduces a novel direction: enhancing video generation efficiency by increasing video autoencoder compression ratios without additional training, thereby reducing overall computational demands.

\section{Method}
In this section, we present \dlfr, a training-free solution for dynamic frame rate control in the latent space. We first establish our theoretical foundation by analyzing the temporal frequency of video signals in both pixel and latent spaces (Sec.\ref{subsec:motivation}). Based on this analysis, we propose our dynamic latent frame rate space (Sec.\ref{subsec:dflr_space}). To realize this design, we propose two key technical components: \raisebox{-1.1pt}{\ding[1.1]{182\relax}} a Dynamic Latent Frame Rate Scheduler that determines optimal frame rates based on content complexity (Sec.\ref{subsec:dflr_Scheduler}), and \raisebox{-1.1pt}{\ding[1.1]{183\relax}} a training-free adaptation mechanism that enables pretrained VAE architectures to process variable frame rate features (Sec.\ref{subsec:dflr_vae}).
Beyond these components, we provide a more straightforward explanation for why our simple yet effective approach can work (Sec.\ref{subsec:discuss}).

\subsection{Motivation}
\label{subsec:motivation}
Traditional frame rate optimization in video processing and compression has predominantly focused on raw video signals \cite{song2001rate,mackin2015study}. Previous studies in video content analysis have demonstrated that video information density exhibits strong temporal non-uniformity. This non-uniformity manifests as significant variations in temporal frequency across different video segments \cite{menon2022vca,papakonstantinou2023content}. 
For instance, in our analysis of the BVI-HFR dataset (Fig.~\ref{fig:video_examples}), fast-motion sequences like ``books'' exhibit 5--8× higher temporal frequency magnitude compared to static scenes like ``flowers'' (see Fig.\ref{fig:analysis_on_video_seg} Top). 

However, with the growing adoption of deep learning for video generation, videos are increasingly mapped into a latent space via an encoder~\cite{fan2024fluid,kuaishou2024,zheng2024open,kong2024hunyuanvideo}. This raises crucial questions: \textit{(i) How do temporal characteristics transfer into the latent space? (ii) Does the latent space preserve the frequency variations observed in the original pixel domain? (iii) Can we apply dynamic frame rates within the latent space?} Although these questions are highly relevant, all of them remain underexplored.

\paragraph{Frequency Analysis of Signals.}

A continuous-time signal can be denoted as $x(t)$. Its frequency spectrum $X(f)$ is obtained via the Fourier transform, and for a  band-limited signal, it is nonzero only up to a maximum frequency $f_{\max}$:
\begin{equation}
X(f) = 0, \quad \forall|f| > f_{\max}.
\end{equation}
Sampling the continuous signal at sampling frequency $F_s$ discretizes $x(t)$ into frames:
\begin{equation}
x[n] = x(nT), \quad T = \frac{1}{F_s}.
\end{equation}
According to the Nyquist-Shannon sampling theorem~\cite{Nyquist_theory,shannon_noise}, $F_s$ must satisfy
\begin{equation}
F_s \geq 2f_{\max}.
\end{equation}
to prevent aliasing. As $f_{\max}$ varies across different segments of the signal, different segments naturally require different frame rates, motivating adaptive frame rate strategies.

\paragraph{Temporal Frequency Analysis of Latent Space.}
Let the video luminance signal be $x(t)$, and its encoder mapping be $\mathcal{E}$. In the latent space, the signal becomes $z(t) = \mathcal{E}(x(t))$ with a  corresponding frequency spectrum $Z(f)$:
\begin{equation}
Z(f) = \int_{-\infty}^{\infty} z(t)e^{-j2\pi ft} dt.
\end{equation}
Note that $\mathcal{E}$ is a complex nonlinear transformation, it can alter the amplitude, phase, and frequency characteristics of a signal or generate new frequency components, leading to changes in the video signal's shape and spectrum.
Therefore, we first analyze the signal in the latent space.
% Hence, the highest frequency in the latent domain, $f'_{\max}$, usually satisfies:
% \begin{equation}
% f'_{\max} \leq g(f_{\max}),
% \end{equation}
% where $g(\cdot)$ denotes the encoder's frequency response function~\cite{smith1997scientist}.

% \begin{figure}[!t]
%     \centering
%     \includegraphics[width=0.99\linewidth]{figures/frequency_analyse.pdf}
%     \caption{Frequency analysis of original video and latent space in temporal dimension. The video segments are 1080p, 60Hz, and 10-second length in BVI-HFR dataset~\cite{mackin2018study}.}
%     \vspace{-0.2in}
%     \label{fig:analysis_on_video_seg}
% \end{figure}

Through empirical analysis on multiple video segments (see Fig.\ref{fig:analysis_on_video_seg}), we observe pronounced frequency variability in both the original video domain and latent domain. High-speed segments—such as rapid camera pans or fast-moving objects—often retain higher temporal frequencies, whereas low-motion segments focus on lower-frequency components. This suggests that adaptive frame rate optimization remains viable in latent space, just as it is in raw pixel space.

In line with the Nyquist-Shannon theorem, when the latent-space sampling rate $F'_s$ meet
\begin{equation}
F'_s \geq 2f'_{\max},
\end{equation}
aliasing can be prevented. The $f'_{\max}$ is its maximum frequency in latent space. Hence, segment-wisely estimating $f'_{\max}$ and adjusting $F'_s$ provides an opportunity to implement a variable frame rate in latent space. 

\begin{figure*}[!tbhp] % h:here 当前位置 % b bottom % t top % p 浮动
    \centering
    \includegraphics[width=0.98\textwidth]{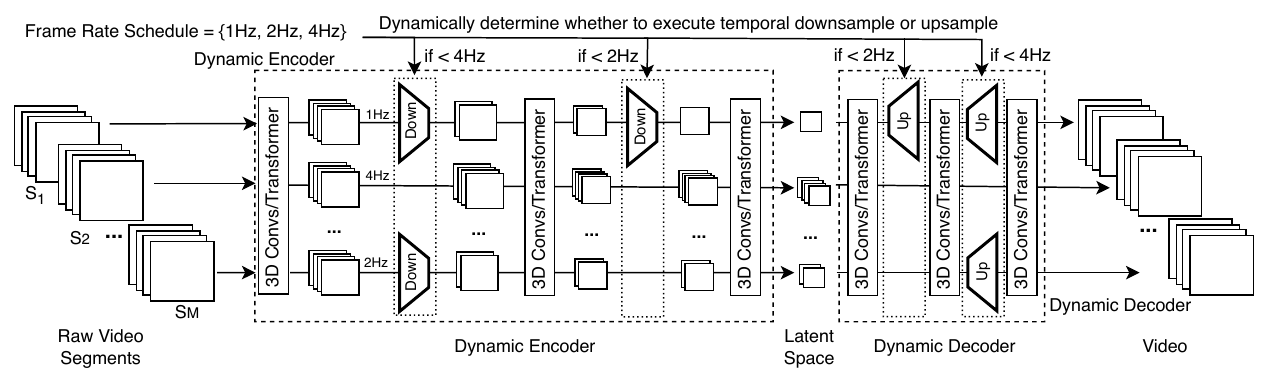} %ims/xx.png
    \vspace{-0.1in}
    \caption{\textbf{Architecture overview of the Dynamic Latent Frame Rate (DLFR) VAE.} The input video is first divided into segments. The dynamic encoder processes these segments through a series of 3D convolution layers interspersed with dynamic downsample operations (Eq.\ref{eq:encoder} in Sec.\ref{subsec:dflr_vae}), where the execution of downsample is determined by the schedule (Sec.\ref{subsec:dflr_Scheduler}). The resulting latent representations maintain varying temporal resolutions according to segment complexity (Sec.\ref{subsec:dflr_space}). The dynamic decoder then reconstructs the video through corresponding upsampling operations (Eq.\ref{eq:decoder} in Sec.\ref{subsec:dflr_vae}), restoring the original frame rate while preserving temporal consistency. Each segment can be processed at different frame rates, enabling content-adaptive temporal compression in latent space.}
    \vspace{-0.1in}
    \label{fig:dynamic_vae}
\end{figure*}

\subsection{Dynamic Frame Rate Latent Space}
\label{subsec:dflr_space}
Building on the above analysis, we propose a dynamic frame rate latent space wherein each video segment can have a distinct frame rate, allocated based on its temporal complexity. Specifically, suppose an input video is divided into $M$ segments $\{S_1,\ldots,S_M\}$, each comprising $N$ frames. For segment $S_i$, its latent representation is $z_i(t)$. The frequency spectrum $Z_i(f)$ is defined as
\begin{equation}
Z_i(f) = \int_{t_i}^{t_i+NT} z_i(t)e^{-j2\pi ft} dt,
\end{equation}
where $T = 1/F_s$ is the sampling interval in the raw video domain.

From empirical observations, certain high-frequency components in $Z_i(f)$ have negligible amplitude and minimal impact on overall fidelity. We thus define an effective maximum frequency $f_{\text{eff},i}$ for each segment, identifying the point where the amplitude remains above a threshold $\epsilon$:
\begin{equation}
f_{\text{eff},i} = \max\{f \mid |Z_i(f)| \geq \epsilon\}.
\end{equation}

By the Nyquist-Shannon principle~\cite{ash2012information}, the corresponding latent-space frame rate for segment $S_i$ can be lowered to
\begin{equation}
F'_{s,i} = 2f_{\text{eff},i}.
\end{equation}
This adaptive sampling ensures each segment maintains only the minimum frame rate necessary to preserve perceptually significant temporal details. To maintain temporal consistency across segment boundaries, we implement a
smooth transition mechanism that gradually adjusts frame
rates between adjacent segments.

\subsection{DLFR Scheduler} 
\label{subsec:dflr_Scheduler} 
While theoretically sound, computing exact frequency spectra for real-time video processing presents significant computational challenges. We address this through a practical approximation strategy that maintains the benefits of dynamic frame rates while ensuring computational efficiency. Our approach discretizes the continuous space of temporal complexities into $N$ distinct levels.
Each level $k \in \{1,\ldots,N\}$ is associated with an effective frequency $f'_{\text{eff},k}$~\footnote{For example, we use \{1, 2, 4\}Hz for the 16 FPS video, which have \{16x, 8x, 4x\} temporal downsample ratio.}. % TODO: the number ablation
If segment $S_i$ falls into complexity class $k$, its latent frame rate becomes
\begin{equation}
F'_{s,i} = 2f'_{\text{eff},k}.
\end{equation}
Formally, we express this as:
\begin{equation}
F'_{s,i} = \sum_{k=1}^N (2f'_{\text{eff},k}) \mathbb{I}_{C_k}(S_i),
\end{equation}
where $\mathbb{I}_{C_k}(S_i)$ is an indicator function that is 1 if $S_i$ belongs to class $C_k$ and 0 otherwise.

\begin{figure}[!tb] % h:here 当前位置 % b bottom % t top % p 浮动
    \centering
    \includegraphics[width=0.9\linewidth]{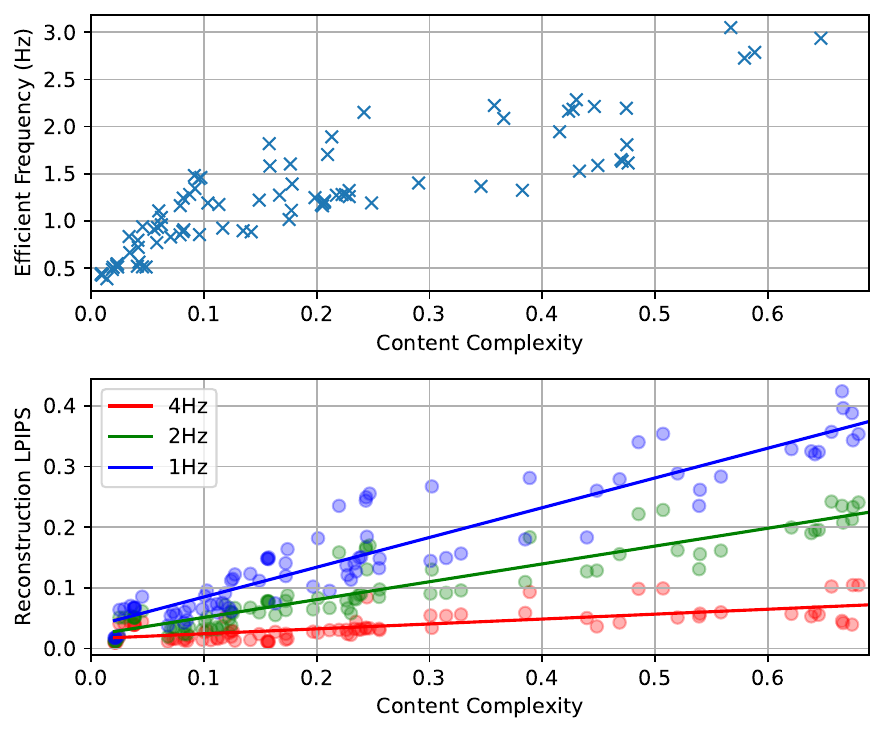} %ims/xx.png
    \caption{Content complexity experiment on HunyuanVideo VAE. The upper figure illustrates the relationship between content complexity and effective frequency, with  $\epsilon=1.8$ used in this experiment. The lower figure demonstrates the alignment between content complexity and reconstruction LPIPS, indicating a strong correlation.}
    \label{im_content_complexity}
\end{figure}

Directly evaluating $\mathbb{I}_{C_k}(S_i)$ from raw or latent signals can still be challenging. 
Instead, we use a practical content complexity metric $C(S_i)$ as a proxy, which considers the SSIM of adjacent frames in a raw video segment: 
\begin{equation}
C(S_i) = \frac{1}{N}\sum_{j=1}^{N-1} (1-\text{SSIM}(x[j],x[j+1])),
\end{equation}
This metric can efficiently distinguish high-motion segments from low-motion ones without explicitly analyzing the latent frequency spectrum. The scheduling logic then maps each segment's metric value to an appropriate complexity class $k$, and hence to a frame rate $F'_{s,i}$. 
As illustrated in Figure~\ref{im_content_complexity}, the content complexity metric exhibits a high correlation with both the effective frequency and the VAE reconstruction performance.
We use this metric and thresholds $Th$ to determine the frame rate:
\begin{equation} 
\label{eq_complexity_threshold}
\mathbb{I}_{C_k}(S_i) = 
\begin{cases} 1, & \text{if}~~~~Th^{down}_{i}<C(S_i)\le Th^{up}_{i} \\
0, & \text{else}. \end{cases} 
\end{equation}

% \[
% \mathbb{I}_{C_k}(S_i) = 
% \begin{cases} 
% 1 & \text{if } S_i \in C_k, \\
% 0 & \text{otherwise}.
% \end{cases}
% \]

% we cannot get the real the temporal space of input video is quantized, 
% 如何评价，提出Content Information

% 具体而言，我们引入了一个动态帧率Scheduler（Dynamic Frame Rate Encoder Scheduler）。该模块首先评估视频每个chunk的信息密度，并动态分配对应的潜空间帧率。

\subsection{Transform Static VAE to Dynamic VAE}
\label{subsec:dflr_vae}

% 如何从原始视频生成动态帧率的潜空间，以及如何从动态潜空间还原回原始视频。我们注意到：以往的视频VAE模型通过在大规模包含各种动作与节奏的视频数据上训练，已经具备了将视频压缩到固定帧率潜空间并还原回原始视频的能力。这一特性启发我们对现有的静态帧率VAE模型进行改造，扩展出动态帧率的潜空间生成能力。我们需要一个运行的过程中动态根据输入scheduler的变化而变化的模型（Encoder模型的输出维度动态变化，Decoder模型的输入维度动态变化），因此我们需要将静态模型转换为一个动态模型。

% 具体来说，我们在VAE的Encoder中插入一个dynamic down下采样器我们在VAE模型中的合适位置根据分配的帧率对VAE中间特征进行动态下采样，从而得到包含不同帧率的chunk特征。在该module之后的VAE模型以相同方式推理具有不同的chunk帧率的features，最终输出的结果就是在dynamic frame rate的latent space。在解码过程中，我们在Deocoder模型中的合适位置根据编码时确定的帧率对中间特征进行上采样，将所有chunk特征恢复到一致的帧率，最终重建出原始视频。我们将其称为DLFR-VAE（Dynamic Latent Frame Rate VAE）

To convert a pretrained static-frame VAE into a dynamic-frame version, we exploit the existing capacity of modern video VAEs, which have learned to compress videos into fixed-frame latent spaces~\cite{chen2024deep,xing2024large,zhu2023designing}. As illustrated in Fig.~\ref{fig:dynamic_vae}, our approach introduces two key modifications to the pretrained VAE: a dynamic downsampling module in the VAE's encoder and a corresponding upsampling module in the decoder. This design allows us to leverage the robust compression capabilities of pretrained VAEs while enabling variable frame rate processing without requiring additional training.

\paragraph{Encoder Modification.} Let the input video be $\{x_1,x_2,\ldots,x_T\}$. A pretrained video VAE encoder $E$ typically processes this input into a latent representation $z$. To support variable frame rates, we introduce a dynamic downsampler at a strategically chosen point in the encoder. Given a frame-rate schedule $\{F'_{s,1},F'_{s,2},\ldots,F'_{s,M}\}$, the downsampler transforms encoder features $h_i$ for each segment $S_i$ into a reduced-rate feature $h'_i$:
\begin{equation}
h'_i = \text{Downsample}(h_i, F'_{s,i}).
\label{eq:encoder}
\end{equation}
These reduced-rate features are then passed through the remaining encoder layers, denoted $E_{\text{post}}$, to yield segment-wise latent codes $\{z_1,\ldots,z_M\}$.

\paragraph{Decoder Modification.} Decoding requires reversing the frame rate changes. A dynamic upsampler is inserted at the corresponding decoder stage. For each segment's latent code $z_i$, the initial decoder layers $D_{\text{pre}}$ produce intermediate features $h''_i$:
\begin{equation}
h''_i = D_{\text{pre}}(z_i).
\end{equation}
The upsampler then restores the original frame rate $F_s$:
\begin{equation}
h'''_i = \text{Upsample}(h''_i, F_s),
\label{eq:decoder}
\end{equation}
after which the remaining decoder layers $D_{\text{post}}$ reconstruct the final segment $\hat{S}_i$. The overall video reconstruction $\hat{V}$ is formed by concatenating $\{\hat{S}_1,\ldots,\hat{S}_M\}$.

Crucially, these modifications allow the pretrained encoder and decoder weights to remain largely unchanged, except for the newly inserted downsampling and upsampling operators. Consequently, \dlfr\ can be deployed as a training-free extension on top of mainstream video VAEs, seamlessly enabling dynamic latent frame rate control.

\subsection{Discussion on \dlfr}
\label{subsec:discuss}
In addition to our information-theoretic formulation, we offer an intuitive explanation for why our \textbf{simple but effective} \dlfr\ can compress latent space with minimal reconstruction loss.
At its core, \dlfr\ dynamically downsamples the pretrained VAE encoder and, in turn, upsamples its decoder—effectively achieving content-dependent spatial-temporal compression without additional training as shown in Fig.\ref{fig:dynamic_vae}.

Pretrained video VAEs~\cite{chen2024deep,xing2024large,zhu2023designing} are typically trained on large-scale datasets that include diverse motion types, ranging from slow to fast, and from sparse to dense. Many training pipelines also involve augmentation techniques (e.g., temporal interpolation) that effectively expose the VAE to slower versions of the same content. Consequently, the VAE develops an internal capacity to represent video content at different temporal scales, but its default latent space is merely configured to operate at a fixed frame rate. \dlfr~can thus be seen as a mechanism that ``reactivates'' this dormant flexibility. By strategically downsampling and upsampling in the encoder and decoder, respectively, \textit{we allow the VAE to adapt to each segment’s temporal complexity, leveraging the latent representational power that was already learned but not previously utilized for frame rate variation.}

% \clearpage
\section{Experiment}

% 为了验证DLFR-VAE的性能，我们将分别使用了HunyuanVideo-VAE与Open-Sora VAE，将其转化为Dynamic VAE，并在不同视频上测试的视频重建性能。

To evaluate the performance of the proposed \textbf{DLFR-VAE} framework, we applied it to two state-of-the-art pretrained VAE models: \textit{HunyuanVideo VAE} \cite{kong2024hunyuanvideo} and \textit{Open-Sora 1.2 VAE} \cite{zheng2024open}. These models were converted into Dynamic VAEs by incorporating our dynamic frame rate mechanism. We then tested their video reconstruction performance on a diverse set of videos.

\subsection{Video Reconstruction}

We conducted extensive experiments to compare the reconstruction quality of videos processed by the original VAE, Static VAE, and Dynamic VAE under different temporal compression ratios (\textbf{CR}). For evaluation, we used three commonly employed metrics: \textbf{SSIM}, \textbf{PSNR}, and \textbf{LPIPS} \cite{zhang2018unreasonable}. Lower LPIPS values indicate better perceptual quality, while higher SSIM and PSNR values signify better structural and pixel-level fidelity.

\begin{table}[!t]
\centering
\caption{Video reconstruction performance on HunyuanVideo VAE. CR indicates the temporal compression ratio. FVMD\cite{liu2024fr} (Frame-wise Video Motion Deviation) is specifically designed to capture the temporal coherence of inter-frame motion. Compared to the traditional rFVD metric, FVMD is more sensitive to unnatural motion transitions and provides a more accurate assessment of video smoothness and temporal consistency, aligning more closely with human perceptual judgments.} 
\label{tab:hunyuan_results}
\resizebox{1.0\linewidth}{!}{
\begin{tabular}{@{}cccccccc@{}}
\toprule
Size/FPS & VAE & CR & SSIM↑ & PSNR↑ & LPIPS↓ & rFVD↓ & FVMD↓ \\ \midrule
240p/15 & Original & 4x & 0.698 & 23.82 & 0.08 & 116 & 623 \\ 
240p/15 & Static & 8x & 0.582 & 22.21 & 0.12 & 268 & 16514 \\ 
240p/15 & Static & 16x & 0.487 & 20.71 & 0.17 & 1244 & 17855 \\ 
240p/15 & Dynamic & 6x & 0.696 & 23.8 & 0.08 & 134 & 687 \\ 
240p/15 & Dynamic & 8x & 0.684 & 23.64 & 0.09 & 278 & 812 \\ 
240p/15 & Dynamic & 12x & 0.631 & 22.77 & 0.12 & 620 & 2345 \\ 
240p/30 & Original & 4x & 0.66 & 22.7 & 0.09 & 108 & 494 \\ 
240p/30 & Static & 8x & 0.59 & 21.7 & 0.11 & 196 & 13192 \\ 
240p/30 & Static & 16x & 0.5 & 20.6 & 0.15 & 1099 & 14777 \\ 
240p/30 & Dynamic & 6x & 0.661 & 22.7 & 0.09 & 127 & 508 \\ 
240p/30 & Dynamic & 8x & 0.655 & 22.6 & 0.09 & 324 & 815 \\ 
240p/30 & Dynamic & 12x & 0.62 & 22 & 0.11 & 571 & 1907 \\ \midrule
540p/15 & Original & 4x & 0.806 & 27.27 & 0.071 & 57 & 331 \\ 
540p/15 & Static & 8x & 0.668 & 24.27 & 0.128 & 218 & 15134 \\ 
540p/15 & Static & 16x & 0.535 & 21.83 & 0.187 & 1281 & 16135 \\ 
540p/15 & Dynamic & 6x & 0.8 & 27.18 & 0.074 & 124 & 392 \\ 
540p/15 & Dynamic & 8x & 0.779 & 26.74 & 0.082 & 170 & 614 \\ 
540p/15 & Dynamic & 12x & 0.695 & 24.51 & 0.119 & 410 & 1766 \\ 
540p/30 & Original & 4x & 0.789 & 26.29 & 0.075 & 60 & 272 \\ 
540p/30 & Static & 8x & 0.682 & 24.05 & 0.118 & 155 & 11784 \\ 
540p/30 & Static & 16x & 0.562 & 21.84 & 0.172 & 1083 & 13222 \\ 
540p/30 & Dynamic & 6x & 0.786 & 26.23 & 0.076 & 131 & 300 \\ 
540p/30 & Dynamic & 8x & 0.773 & 26 & 0.082 & 156 & 526 \\ 
540p/30 & Dynamic & 12x & 0.712 & 24.2 & 0.109 & 485 & 1825 \\ \midrule 
720p/15 & Original & 4x & 0.971 & 38.95 & 0.041 & 8 & 201 \\ 
720p/15 & Static & 8x & 0.806 & 30.71 & 0.109 & 163 & 15180 \\ 
720p/15 & Static & 16x & 0.65 & 25.32 & 0.173 & 1257 & 16345 \\ 
720p/15 & Dynamic & 6x & 0.962 & 37.97 & 0.045 & 34 & 291 \\ 
720p/15 & Dynamic & 8x & 0.927 & 35.47 & 0.057 & 132 & 463 \\ 
720p/15 & Dynamic & 12x & 0.834 & 30.95 & 0.095 & 304 & 1687 \\ 
720p/30 & Original & 4x & 0.97 & 38.95 & 0.041 & 8 & 200 \\ 
720p/30 & Static & 8x & 0.832 & 30.07 & 0.097 & 189 & 15180 \\ 
720p/30 & Static & 16x & 0.684 & 25.68 & 0.162 & 1257 & 16348 \\ 
720p/30 & Dynamic & 6x & 0.964 & 38.07 & 0.04 & 94 & 291 \\ 
720p/30 & Dynamic & 8x & 0.944 & 36 & 0.049 & 156 & 411 \\ 
720p/30 & Dynamic & 12x & 0.867 & 31.32 & 0.081 & 505 & 1687 \\ \bottomrule 
\end{tabular}
}
\vspace{-0.2in}
\end{table}

\begin{table}[!t]
\centering
\caption{Video reconstruction performance on Open-Sora VAE. CR indicates the temporal compression ratio.} 
\label{tab:Open-Sora_results}
\resizebox{0.9\linewidth}{!}{
\begin{tabular}{@{}cccccccc@{}}
\toprule
Size/FPS & VAE & CR & SSIM↑ & PSNR↑ & LPIPS↓ & rFVD↓ & FVMD↓ \\ \midrule 
540p/15 & Original & 4x & 0.856 & 29.62 & 0.199 & 435 & 899 \\ 
540p/15 & Static & 8x & 0.746 & 27.31 & 0.245 & 1103 & 14311 \\ 
540p/15 & Static & 16x & 0.659 & 25.71 & 0.27 & 2164 & 15311 \\ 
540p/15 & Dynamic & 6x & 0.849 & 29.39 & 0.199 & 516 & 998 \\ 
540p/15 & Dynamic & 8x & 0.822 & 28.8 & 0.21 & 862 & 1085 \\ 
540p/15 & Dynamic & 12x & 0.762 & 27.44 & 0.229 & 1352 & 3577 \\ 
540p/30 & Original & 4x & 0.866 & 29.91 & 0.181 & 311 & 789 \\ 
540p/30 & Static & 8x & 0.781 & 27.99 & 0.229 & 986 & 13299 \\ 
540p/30 & Static & 16x & 0.702 & 26.23 & 0.257 & 1877 & 14311 \\ 
540p/30 & Dynamic & 6x & 0.864 & 29.9 & 0.177 & 340 & 823 \\ 
540p/30 & Dynamic & 8x & 0.849 & 29.3 & 0.188 & 499 & 1001 \\ 
540p/30 & Dynamic & 12x & 0.793 & 27.92 & 0.213 & 1277 & 3571 \\ \midrule  
720p/15 & Original & 4x & 0.884 & 30.73 & 0.179 & 305 & 799 \\ 
720p/15 & Static & 8x & 0.779 & 28.33 & 0.221 & 1003 & 13656 \\ 
720p/15 & Static & 16x & 0.692 & 26.6 & 0.244 & 2071 & 13699 \\ 
720p/15 & Dynamic & 6x & 0.877 & 30.42 & 0.182 & 551 & 863 \\ 
720p/15 & Dynamic & 8x & 0.85 & 29.78 & 0.192 & 737 & 1070 \\ 
720p/15 & Dynamic & 12x & 0.792 & 28.35 & 0.211 & 1232 & 2946 \\ 
720p/30 & Original & 4x & 0.897 & 31.22 & 0.159 & 320 & 534 \\ 
720p/30 & Static & 8x & 0.818 & 29.22 & 0.207 & 940 & 8848 \\ 
720p/30 & Static & 16x & 0.739 & 27.3 & 0.234 & 2131 & 8849 \\ 
720p/30 & Dynamic & 6x & 0.896 & 31.16 & 0.158 & 519 & 654 \\ 
720p/30 & Dynamic & 8x & 0.882 & 30.51 & 0.168 & 925 & 659 \\ 
720p/30 & Dynamic & 12x & 0.828 & 29.09 & 0.194 & 1345 & 1660 \\ \bottomrule 
\end{tabular}
}
\vspace{-0.2in}
\end{table}

% 为了测试不同的动态特性的视频的影响，我们选取了BVI-HFR dataset，这个数据集涵盖了多种场景类型和运动模式，包括运动模糊和动态纹理。例如，有运动场景、静态背景下的快速运动物体等。我们分别测试了分辨率为540p和720p的视频在15fps、30fps下的性能。

% 原始的VAE的temporal压缩率为4x，为了将Static VAE转换为Dynamic VAE，我们在Encoder中原本的前两个temporal strided conv后，分别加入一个Dynamic downsample operator，在Decoder中原本的最后两个temporal strided conv后，分别加入一个Dynamic upsample operator。
% 为了验证动态根据输入信息进行动态判断下采样比例的好处，我们还设置了一个插入静态下采样和上采样算子的VAE对比对象，它们的temporal压缩率分别为8x和16x。

% 在Table 1和Table 2中，我们展示了将Threshold设置为平均temporal compression ratio为8x和12x的位置时的性能（Section 4.3中有关于threshold的详细ablation实验）。从实验结果可以有以下发现：1. 相比于原始的VAE，直接通过插入Static downsample和upsample operations会大幅度降低VAE的视频重建性能。2. Dynamic VAE相比Static VAE可以大幅度提升视频重建性能。在12x压缩率的情况下甚至比static VAE在8x压缩率的情况下有更好的性能。 3. Dynamic VAE在不同模型、不同分辨率、不同原始视频帧率下均有不错的表现。这证明了DLFR-VAE方法的通用性。

To test the effects of dynamic characteristics in videos, we used the \textbf{BVI-HFR dataset} \cite{mackin2018study}, which includes a variety of scene types and motion patterns, such as dynamic textures and fast-moving objects against static backgrounds. We evaluated videos at two resolutions (\textbf{540p} and \textbf{720p}) and two frame rates (\textbf{15fps} and \textbf{30fps}).

For the original VAE, the temporal compression ratio is 4x. To convert the Static VAE into a Dynamic VAE, we modified the encoder and decoder by introducing Dynamic Downsampling and Dynamic Upsampling operators. For HunyuanVideo VAE, the encoder's first two temporal strided convolution layers were augmented with a Dynamic Downsampling operator, while the decoder's last two temporal strided convolution layers were enhanced with a Dynamic Upsampling operator.
For Open-Sora VAE, we add the dynamic downsampling operators after the 2D VAE encoder, and the dynamic upsampling operators before the 2D VAE decoder. Both are a bilinear sample.

The results in Tables~\ref{tab:hunyuan_results} and~\ref{tab:Open-Sora_results} highlight the following key findings: 1. \textbf{Static VAE significantly reduces reconstruction quality:} Compared to the original VAE, inserting static downsampling and upsampling operators at higher compression ratios (\textbf{8x} and \textbf{16x}) leads to a noticeable drop in SSIM, PSNR, and LPIPS performance. This demonstrates the limitations of static compression in preserving video fidelity. 2. \textbf{Dynamic VAE outperforms Static VAE:} The proposed DLFR-VAE framework consistently achieves higher reconstruction quality than its static counterpart across all resolutions, frame rates, and compression ratios. 
At a compression ratio of \textbf{6x}, the performance of dynamic VAE is comparable to the Original VAE. At a compression ratio of \textbf{12x}, the Dynamic VAE even surpasses the Static VAE at \textbf{8x} in terms of reconstruction metrics, demonstrating the effectiveness of the adaptive approach in optimizing information retention. 3. \textbf{Generalizability of DLFR-VAE:} The performance gains of Dynamic VAE hold across different pretrained models, resolutions, and frame rates. This highlights the generality and robustness of the DLFR-VAE framework, making it suitable for a wide range of video generation tasks.

\subsection{Visualization}

\begin{figure*}[!tb] % h:here 当前位置 % b bottom % t top % p 浮动
    \centering
    \includegraphics[width=0.95\linewidth]{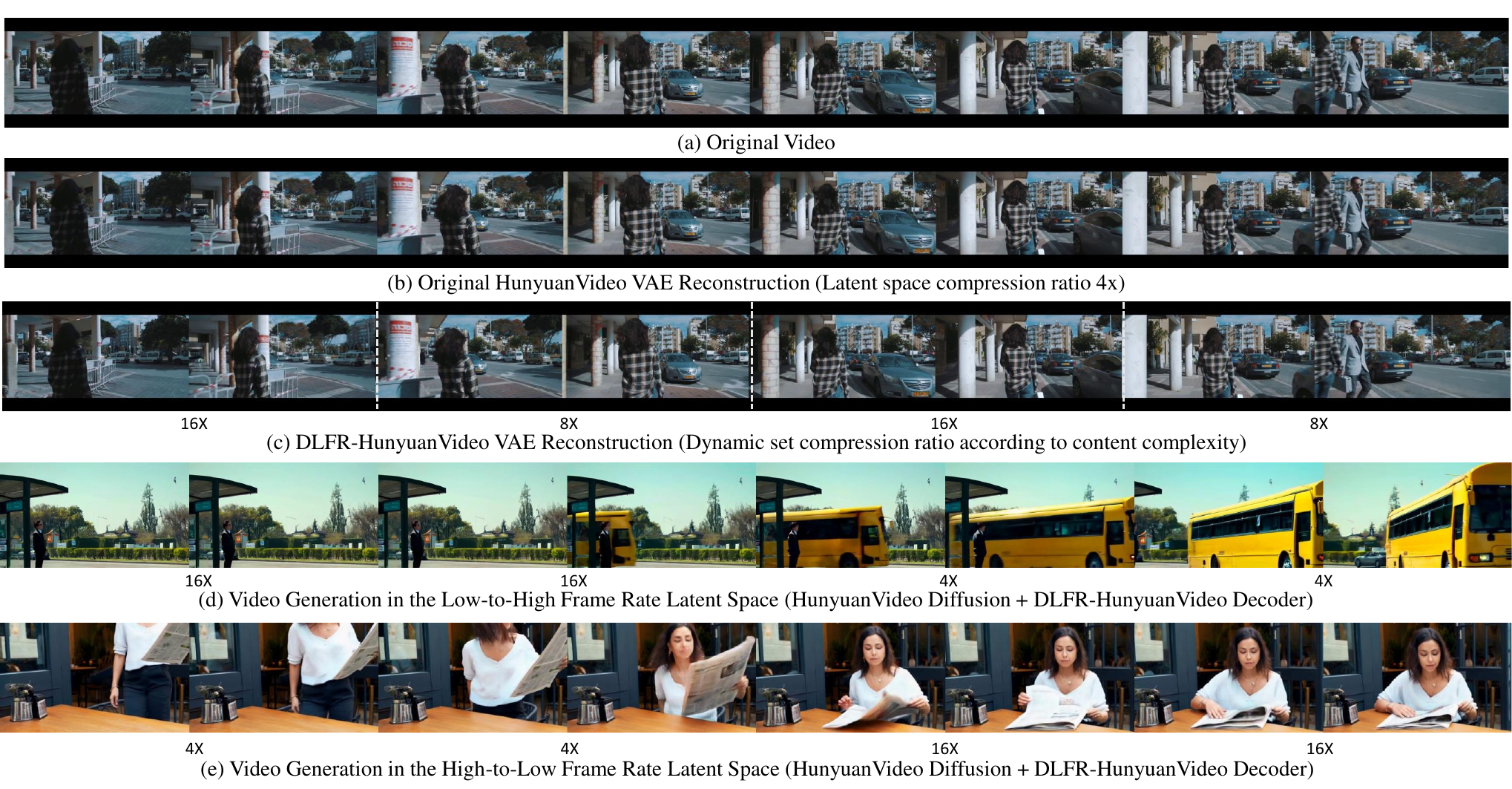} %ims/xx.png
    \caption{Comparison of the (a) original video, (b) the reconstruction result using the original HunyuanVideo VAE, and (c) the reconstruction result using our proposed DLFR-VAE. The figure illustrates the effectiveness of our dynamic frame rate adaptation in preserving video quality while reducing computational overhead. (d,e) The generated video in the dynamic latent space using the prompt: \textit{Realistic style. A man stands at a quiet bus stop on a sunny afternoon. Then, a bright yellow bus approaches.} and \textit{A woman strolls into a café and approaches a wooden table. She picks up a newspaper and starts reading it}.}
    \label{im_visualization}
\vspace{-0.2in}
\end{figure*}

% 如图所示，我们对DLFR-VAE的重建结果进行展示。可以发现，在摄像机视角快速变化的场景使用了较低的temporal compression ratio，也就是较高的Frame rate；在摄像机视角相对静止位置，使用较高的compression ratio，也就是较低的Frame rate。从主观视觉感受来说，视频质量在整体语意表达和xx，但在细节上有小幅下降。仔细观察可以发现，尤其是16x的压缩率，有一定的视频质量损失，尤其是在一些局部快速变化的物体上，会出现唯影等问题。但需要注意的是，我们是一个Training free的方法。未来可通过VAE finetune或者重头训练等方式缓解，或者通过对画面不同位置使用不同的frame rate来解决局部与整体差异的问题。

To visually demonstrate the effectiveness of our DLFR-VAE framework, we present a comparison between the original video, the reconstruction result from the original HunyuanVideo VAE, and the reconstruction result from DLFR-HunyuanVideo VAE. As shown in Figure~\ref{im_visualization}(a,b,c), our method dynamically adjusts the temporal compression ratio based on the complexity of the video content. Specifically, in scenes with rapid camera movements or high motion, DLFR-VAE employs a lower temporal compression ratio (i.e., a higher frame rate) to preserve temporal details. Conversely, in scenes with relatively static camera views or low motion, a higher compression ratio (i.e., a lower frame rate) is used to reduce computational overhead.

From a subjective visual perspective, the overall semantic content and motion coherence of the reconstructed video are well-preserved. However, there is a slight degradation in fine-grained details, particularly in regions with rapid local motion. For instance, at a compression ratio of 16x, some artifacts such as motion blur or ghosting may appear in areas with fast-moving objects.

It is important to note that DLFR-VAE is a training-free approach, meaning it does not require additional training or fine-tuning of the underlying VAE model. While this makes our method highly efficient and easy to integrate with existing systems \cite{kong2024hunyuanvideo, zheng2024open}, the observed quality degradation in high-compression scenarios suggests potential areas for future improvement. For example, fine-tuning the VAE or training it from scratch with dynamic frame rate adaptation could help mitigate these artifacts. Additionally, a more sophisticated approach could involve applying different frame rates to different regions of the video frame, addressing the trade-off between global and local motion preservation.

\begin{figure}[!tb] % h:here 当前位置 % b bottom % t top % p 浮动
    \centering
    \includegraphics[width=0.9\linewidth]{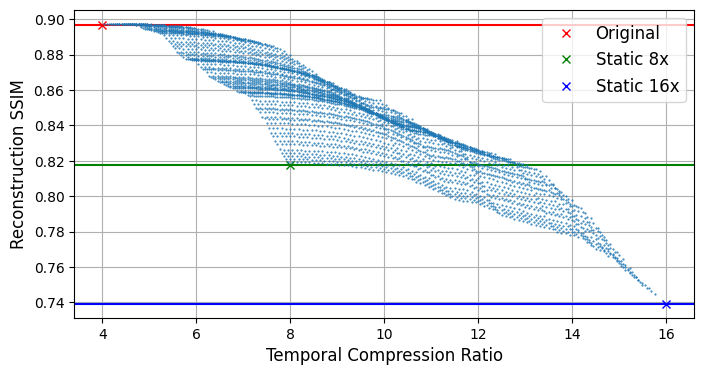} %ims/xx.png
\vspace{-0.2in}
    \caption{The influence of different threshold settings on Open-Sora 720p 30fps videos. The threshold grid is made to cover the whole range of content complexity.}
    \label{im_ablation_threshold}
\vspace{-0.2in}
\end{figure}

\subsection{Ablation Study on Threshold}

% \textbf{Dynamic VAE Encoder Structure}

% \textbf{Dynamic VAE Decoder Structure}

% \subsubsection{The Influence of Threshold} 

% 公式12中的threshold会影响结果，为了研究不同的threshold设置对结果的影响，我们将下采样的threshold做了一个grid，并测试了不同threshold下的reconstruction SSIM。结果如图5所示。从图中可以看出，不同的threshold下，Dynamic VAE的性能都比Static的性能要高；同时，dynamic VAE也有清晰明显的帕累托最优曲线，这也显示了dynamic VAE的通用性，可以满足不同压缩率的需求。

The threshold parameter, introduced in Equation~\ref{eq_complexity_threshold}, plays a critical role in determining the performance of the proposed DLFR-VAE. To better understand its impact, we conducted a grid search over different threshold values and evaluated the reconstruction performance. The results are visualized in Figure~\ref{im_ablation_threshold}.

From Figure~\ref{im_ablation_threshold}, the following observations can be drawn: 
1. Across all tested threshold values, the performance of the \textbf{Dynamic VAE} consistently surpasses that of the \textbf{Static VAE}. This demonstrates the robustness and adaptability of the Dynamic VAE framework, as it effectively balances compression and reconstruction quality.
2. The presence of a well-defined Pareto curve \cite{blanchet2022generalized} underscores the versatility of the Dynamic VAE across varying compression levels. This generalizability suggests that the model is capable of adapting to diverse use cases, from low-bitrate streaming to high-fidelity video reconstruction.

% \textbf{Content Complexity Metric}. 
% We tried different kinds of metric to compute the difference between the frames, including MSE SSIM PSNR.

\subsection{Video Generation in Dynamic Latent Space}

% 我们尝试了在不经过训练的情况下，在Dynamic Frame Rate Latent Space中进行视频，由于当前图像生成网络几乎都是DiT架构，我们只需要调整positional encoding即可实现不同的frame rate的token在同时生成。在生成之前，我们需要人工设置每一段生成的帧率，我们使用了两种帧率设置，第一种是前面高帧率后面低帧率，第二种是前面低帧率后面高帧率。我们在图8中展示了生成的结果。结果显示，即使不经过训练，DiT神经网络可以在Dynamic Frame Rate Latent Space中生成正常的视频。

% 当然，本文的主要内容是如何构建一个Dynamic的Latent Space，而不是探索如何在这个空间中进行生成，这里只是展示一个可能性。我们认为在未来，可以根据用户的prompt，由神经网络自动判断每一段视频应该生成多少帧率的视频。或者从头训练在Dynamic空间中的视频生成模型。

% \subsection{}

In this section, we explore the feasibility of generating videos directly within a \textit{Dynamic Frame Rate Latent Space} without any additional training. Given that contemporary image and video generation models are predominantly based on the Diffusion Transformer (DiT) architecture \cite{peebles2023scalable,esser2024scaling}, adapting these models to work in a dynamic latent space requires only minor adjustments. 

% To get the latent frame rates for different temporal segments of the generated video, we first generate a video with a very small number of denoising steps (such as 3 steps) and then we using the DLFR scheduler to determine the frame ratios of video segments. Next we use the frame rate configuration to generate the videos in the dynamic frame rate latent space. Finally, we use our DLFR-VAE decoder to convert the latents into video.

To obtain the latent frame rates for different temporal segments of the generated video, we follow a multi-step process. First, we generate a preliminary video using a minimal number of denoising steps (e.g., 5 steps), which has a 10\% computation cost of generating the video with 50 steps. Although this preliminary video has very low quality due to the limited number of steps, it still provides a clear indication of whether each video segment contains fast or slow motion. Based on this information, we then use the DLFR scheduler to assess the content complexity of each segment and determine the appropriate frame rate ratios. With these frame rate configurations, we proceed to generate the final videos in the dynamic frame rate latent space. To handle latent frames with varying frame rates, where a frame can represent either a short or long duration of time, we adjust the positional embedding for RoPE (Rotary Position Embedding) in DiT. Because tokens in the compressed latent space that each token represents a longer duration, we resample the cosine and sine parameters for RoPE~\footnote{See Appendix for detail of the positional embedding parameter generation.}. Finally, we apply our DLFR-VAE decoder to convert the latent representations into the final video output.

\begin{table}[tb]
\centering
\caption{Comparing the quality of video generation in raw latent space and our DLFR latent space. The end-to-end latency includes the preliminary video generation, dynamic frame rate schedule and the video generation in the compressed latent space.}
\label{tab:gen_result}
\resizebox{1.0\linewidth}{!}{
\begin{tabular}{@{}cccccc@{}}
\toprule
     & CLIPT  & CLIPSIM & VQA   & FLICKER & Latency (s) \\ \midrule
raw  & 0.9991 & 0.1869  & 97.56 & 0.9781  & 186         \\
ours & 0.9994 & 0.1874  & 97.44 & 0.9852  & 95          \\ \bottomrule
\end{tabular}
}
\end{table}

In Table~\ref{tab:gen_result}, we present the results of generating videos at 480p resolution with 97 frames, based on a subset of prompts collected from VBench \cite{huang2024vbench} with 21 prompts. To evaluate the alignment between text and video, we use CLIPSIM \cite{wu2021godiva}. For assessing temporal consistency, we employ CLIP-Temp \cite{esser2023structure}. Additionally, we use VQA \cite{wu2022fastvqa} to evaluate the aesthetic quality of the generated videos. We also measure temporal consistency, focusing on local and high-frequency details, using the temporal flickering score \cite{huang2024vbench}. Our observations indicate that our approach achieves scores comparable to raw video generation while offering a 2x speedup.

In Figure~\ref{im_visualization}(d,e), we show two videos with 1. High-to-Low Latent Frame Rate: Higher frame rates were applied to the initial temporal segments, followed by lower frame rates in subsequent segments.
2. Low-to-High Latent Frame Rate: Lower frame rates were applied to the earlier segments, gradually increasing to higher frame rates in later segments. Remarkably, even without retraining, the DiT-based neural network was able to produce coherent and visually plausible videos in the dynamic latent space. 

% Since our method focuses on transforming static VAE into dynamic VAE, we did not explore the generation process in depth. This experiment reveals several interesting possibilities for future research: 1. Frame Rate Prediction: Future models could leverage prompts or contextual information to automatically infer the appropriate frame rate for each segment. 2. Training in Dynamic Latent Space: While our approach demonstrates that pretrained generative models can produce videos in a dynamic latent space, training models within this space could yield optimized results. 

\section{Conclusion}

In this paper, we introduced DLFR-VAE, a training-free framework for dynamic latent frame rate adaptation in video generation. By dynamically adjusting the frame rate based on content complexity, DLFR-VAE significantly reduces the number of elements in latent space.
Our experiments demonstrate its effectiveness across various resolutions and frame rates, showcasing its potential as a plug-and-play solution for existing video generation models. 

% \clearpage
% \section*{Impact Statement}


% \section{Impact Statement}

\textbf{Acknowledgment}:
We would like to express our gratitude to Xuefei Ning for her invaluable guidance and support throughout this research endeavor. We are also grateful to the NICS-efc Lab and Infinigence AI for their substantial support.

\clearpage
\bibliography{main}
\bibliographystyle{icml2025}

\clearpage
%%%%%%%%%%%%%%%%%%%%%%%%%%%%%%%%%%%%%%%%%%%%%%%%%%%%%%%%%%%%%%%%%%%%%%%%%%%%%%%
%%%%%%%%%%%%%%%%%%%%%%%%%%%%%%%%%%%%%%%%%%%%%%%%%%%%%%%%%%%%%%%%%%%%%%%%%%%%%%%
% APPENDIX
%%%%%%%%%%%%%%%%%%%%%%%%%%%%%%%%%%%%%%%%%%%%%%%%%%%%%%%%%%%%%%%%%%%%%%%%%%%%%%%
%%%%%%%%%%%%%%%%%%%%%%%%%%%%%%%%%%%%%%%%%%%%%%%%%%%%%%%%%%%%%%%%%%%%%%%%%%%%%%%
\newpage
\appendix
% \onecolumn

\section{Limitations and Future Directions}

Although the current results are promising, there are also limitations. Firstly, although the generated videos using the compressed latent space exhibit good visual quality and high quality scores, they differ from those generated in the original latent space.  Additionally, the lack of retraining prevents the generative model from fully exploiting the advantages of a dynamic latent space. Future research could address these limitations by: 1. Developing automated frame rate schedulers integrated with the generative process. 2. Designing new positional encoding mechanisms tailored to dynamic latent spaces. 3. Training generative models end-to-end in such spaces to maximize efficiency and performance.

\section{Impact Statement}

This paper presents work whose goal is to advance the field of Machine Learning, particularly in the domain of video generation. By introducing DLFR-VAE (Dynamic Latent Frame Rate Variational Auto Encoder), we propose a training-free framework that dynamically adjusts the latent frame rate based on video content complexity, significantly reducing computational overhead while maintaining high reconstruction quality. This innovation has the potential to make video generation more efficient and scalable, enabling longer and higher-resolution video synthesis with reduced computational resources.

The broader impact of this work includes potential applications in various fields such as entertainment, education, and virtual reality, where efficient video generation is crucial. By lowering the computational barriers, DLFR-VAE could democratize access to advanced video generation technologies, allowing smaller organizations and researchers with limited resources to leverage state-of-the-art video synthesis tools.

However, as with any generative technology, there are ethical considerations to be mindful of. The ability to generate high-quality videos efficiently could be misused for creating deepfakes or other forms of misinformation. It is important for the community to develop robust detection mechanisms and ethical guidelines to mitigate such risks. Additionally, the environmental impact of reduced computational requirements could be positive, as it may lead to lower energy consumption in data centers. While this work primarily aims to advance the technical capabilities of video generation, we encourage ongoing discussions around its ethical implications and societal consequences to ensure that the technology is used responsibly and for the benefit of society.

\section{Design Details of Dynamic Upsampler / Downsampler}
We illustrate the design based on the 3D VAE with the 884 architecture adopted by HunyunVideo. The encoder and decoder of this VAE consist of four down blocks / up blocks and a mid block, respectively.

\textbf{In the encoder}, temporal downsampling is achieved by the DownsampleCausal3D modules within certain down blocks, which apply convolutions with stride $> 1$ along the temporal dimension. 

\textbf{In the decoder}, temporal upsampling is not performed via transposed convolution. Instead, each up block restores the temporal sequence length using interpolation (e.g., nearest-neighbor), followed by convolutions with stride $= 1$ for feature fusion and smoothing.

DLFR operates \textbf{only} along the temporal dimension. The design and positional selection of the upsampler/downsampler are as follows:

\paragraph{Temporal Downsampling in the Encoder:}
We modify the temporal stride of the DownsampleCausal3D modules in selected down blocks — for example, changing the stride from $(1,2,2)$ to $(2,2,2)$, or from $(2,2,2)$ to $(4,2,2)$ — where $(T, H, W)$ denotes temporal, height, and width dimensions, respectively. This enables a further increase in the temporal downsampling ratio.

\textbf{Temporal Downsampling Positional Selection:} Among the four down blocks, only the first three (closer to the input) contain DownsampleCausal3D modules, thus offering three selectable positions. For \textbf{2$\times$ downsampling}, one of the three down blocks is selected and its stride is modified; for \textbf{4$\times$ downsampling}, two of them are selected for stride adjustment.

\paragraph{Temporal Upsampling in the Decoder:}
Each up block contains three ResNet blocks and one interpolation module. To perform upsampling, we insert an additional nearest-neighbor interpolation module either \textit{before or after} a selected ResNet block to increase the temporal resolution.

\textbf{Temporal Upsampling Positional Selection:} Across the four up blocks, there are 12 ResNet blocks. Considering both pre- and post-insertion for each, we obtain 16 candidate positions. For \textbf{2$\times$ upsampling}: one interpolation location is selected; for \textbf{4$\times$ upsampling}: two positions are selected.

\paragraph{Optimal Combination Selection:}
To determine the best combinations of upsampling and downsampling positions, we exhaustively evaluate all possible configurations by constructing corresponding VAE models. For \textbf{2\,Hz} frame rate: there are 48 possible combinations; for \textbf{4\,Hz} frame rate: there are 546 possible combinations.These are tested on 500 videos at 15\,Hz and 540p resolution. Evaluation metrics include \textbf{SSIM}, \textbf{PSNR}, and \textbf{LPIPS}.
\vspace{-1.5em} 
\begin{figure}[H]
    \centering
    \hspace*{-1cm}
    \includegraphics[width=1.1\linewidth]{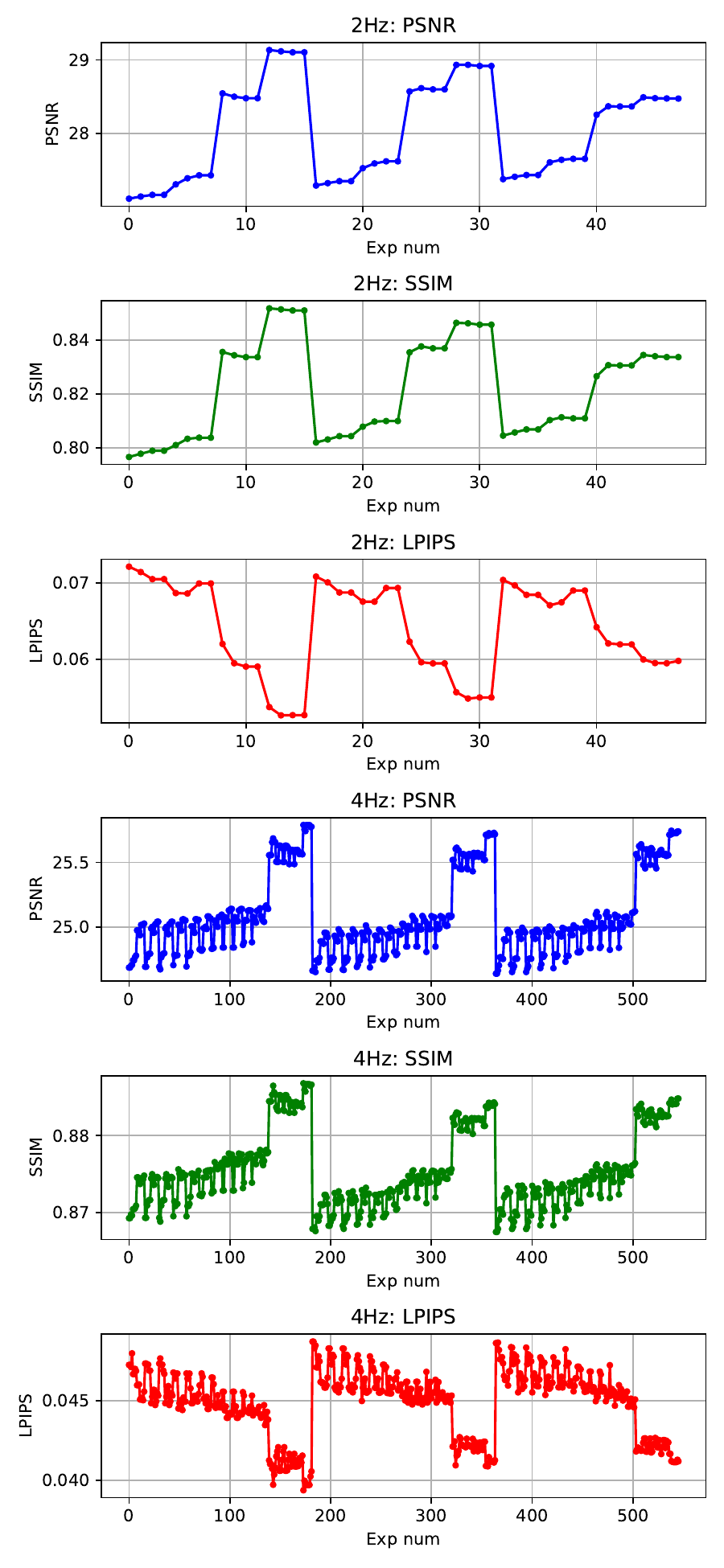}
    \caption{VAE reconstruction metrics (SSIM, PSNR, LPIPS) under different frame rates and combinations.}
    \label{fig:vae_metrics}
\end{figure}
\paragraph{Results:}
The results (see Figure~\ref{fig:vae_metrics}) indicate that the three evaluation metrics (SSIM, PSNR, LPIPS) follow a highly consistent pattern across all configuration combinations. Specifically, configurations achieving the highest SSIM and PSNR values also tend to yield the lowest LPIPS scores. Notably, the most effective upsampling and downsampling positions are predominantly located near the input end of the encoder and the output end of the decoder.

Based on this analysis, we fix the optimal position combinations under each frame rate setting for use in subsequent experiments and system deployment.

\section{Video Generation with Modified RoPE}

For video generation on the dynamic frame rate latent space, we modify the RoPE in of the diffusion transformer (DiT). To handle latent frames with varying frame rates, where a frame can represent either a short or long duration of time, we adjust the positional embedding for RoPE (Rotary Position Embedding) in DiT. Specifically, for tokens in the compressed latent space that each token represents a longer duration, we reduce the frequency of the RoPE parameters to generate the cosine and sine components.
`
\subsection{Concept of RoPE}

The RoPE (Rotary Position Embedding) mechanism is a technique used in transformer-based models~\cite{su2024roformer}, including those applied in diffusion models like Diffusion Transformer. We first provides a detailed introduction to the RoPE mechanism in the context of Diffusion Transformer.

In transformer models, position embeddings are crucial for the model to understand the order of input elements (e.g., tokens in a sequence). Traditional position embeddings (like those in the original Transformer model) use fixed vectors to represent positions, which can be inefficient for long sequences and may not generalize well to sequences of different lengths. RoPE addresses these limitations by introducing a more dynamic way of encoding positions.

RoPE represents positions using a combination of sine and cosine functions. For a token at position $m$ with embedding $\mathbf{x}_m \in \mathbb{R}^D$, RoPE applies rotation matrices $\mathbf{R}_m$ to queries ($\mathbf{q}_m$) and keys ($\mathbf{k}_n$) in the attention mechanism:

\begin{equation}
    \mathbf{q}_m = \mathbf{R}_m \mathbf{W}_q \mathbf{x}_m, \quad \mathbf{k}_n = \mathbf{R}_n \mathbf{W}_k \mathbf{x}_n
\end{equation}

where $\mathbf{R}_m$ is defined block-diagonally for each dimension pair $(2i, 2i+1)$:

\begin{equation}
    \mathbf{R}_m = \bigoplus_{i=1}^{D/2} \begin{pmatrix}
        \cos m\theta_i & -\sin m\theta_i \\
        \sin m\theta_i & \cos m\theta_i
    \end{pmatrix}, \quad \theta_i = 10000^{-2i/D}
\end{equation}

The attention score $A_{m,n}$ naturally encodes relative positions:

\begin{equation}
    A_{m,n} = (\mathbf{R}_m \mathbf{q}_m)^\top (\mathbf{R}_n \mathbf{k}_n)
    \label{eq_raw_rope}
\end{equation}
The frequency of the RoPE is ${10000^{-2i/D}}$.

\subsection{Resample RoPE Parameters for DLFR}

\begin{figure}[tb] % h:here 当前位置 % b bottom % t top % p 浮动
    \centering
    \includegraphics[width=0.47\textwidth]{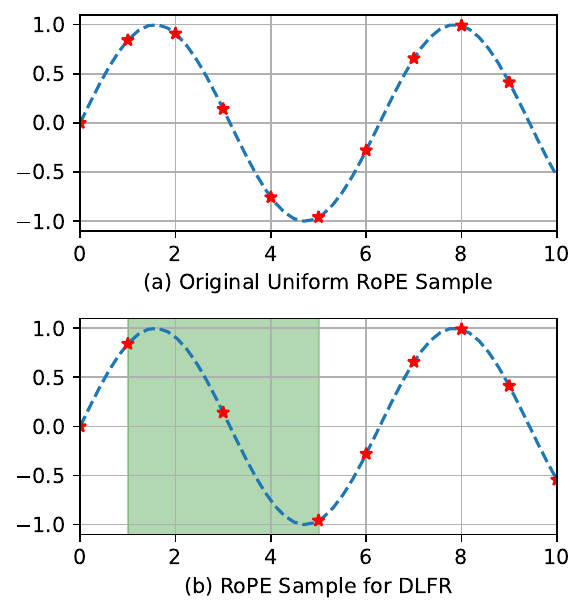} %ims/xx.png
    \caption{Illustration of resampling for DLFR. (a) represents the original RoPE sampling method, while (b) depicts the resampling method adapted for DLFR. The green areas highlight the extended positional relationships for tokens that represent longer time spans.}
\label{im_resample}
\end{figure}

When a latent token corresponds to a longer time span, the RoPE parameters should be adjusted. This is because such tokens encode positional relationships over longer temporal intervals compared to adjacent frames.

To address this, we replace the $m$ in Equation~\ref{eq_raw_rope} with $P_m$ based on the frame rates of different tokens. For a token with longer time span, its positional distance to neighboring tokens increases proportionally. Using this relationship, we compute the new RoPE parameters as follows:

In this way, we generate the new RoPE parameters:

\begin{equation}
    \mathbf{R}_m = \bigoplus_{i=1}^{D/2} \begin{pmatrix}
        \cos P_{m}\theta_i & -\sin P_{m}\theta_i \\
        \sin P_{m}\theta_i & \cos P_{m}\theta_i
    \end{pmatrix}, \quad \theta_i = 10000^{-2i/D}
\end{equation}

As shown in Figure~\ref{im_resample}, the interval of different RoPE sample points is different according to the frame rate.

\subsection{Speedup of generating videos in compressed space}

\begin{table}[!tb]
\centering
\caption{Latency of one diffusion step under different latent space on HunyuanVideo Diffusion model.}
\label{table_gen_latency}
\resizebox{0.9\linewidth}{!}{
\begin{tabular}{@{}cccccc@{}}
\toprule
                      &         & Original & DLFR  & DLFR  & DLFR \\ \midrule
                      & CR      & 4x       & 6x    & 8x    & 12x  \\ \midrule
\multirow{2}{*}{720p} & latency & 59.90    & 29.36 & 19.26 & 9.58 \\
                      & speedup & 1.00     & 2.04  & 3.11  & 6.25 \\ \midrule
\multirow{2}{*}{540p} & latency & 24.93    & 11.61 & 7.73  & 4.05 \\
                      & speedup & 1.00     & 2.15  & 3.23  & 6.16 \\ \bottomrule
\end{tabular}
}
\vspace{-0.2in}
\end{table}

% 我们测试了xxx，从表中可以观察到，DLFR可以显著降低开销，在CR=6x，平均可以获得2x的speedup，在CR=8x，xxx，在CR=12x，xxx。速度提升主要来源于diffusion的token数量的降低，由于Transformer的Attention的计算量是随着token数量平方级别，因此可以显著降低计算量。

We evaluated the latency of denoising step of different compression ratio.
From Table~\ref{table_gen_latency}, we observe that denoising in the compressed latent space can significantly reduces the generation cost. At CR=6x, it achieves an average speedup of 2x for one step of denoising. At CR=12x, DLFR achieves an speedup of more than 6x. These improvements primarily stem from the reduction in the number of tokens processed by the diffusion model. Since the computational complexity of the Transformer's attention mechanism scales quadratically with the number of tokens, reducing the token count leads to substantial computational savings.

% \section{Visualization}

% \subsection{Open-Sora Reconstruction Visualization}
% We provides more 

% \subsection{}

\end{document}